\documentclass[10pt,conference,letterpaper]{IEEEtran}
\IEEEoverridecommandlockouts
\usepackage{cite}
\usepackage{amsmath,amssymb,amsfonts}
\usepackage{algorithmic}
\usepackage{graphicx}
\usepackage{textcomp}
\usepackage{xcolor}
\usepackage{multicol}
\usepackage{multirow}
\usepackage{url}

\def\BibTeX{{\rm B\kern-.05em{\sc i\kern-.025em b}\kern-.08em
    T\kern-.1667em\lower.7ex\hbox{E}\kern-.125emX}}
\begin{document}

\title{Input Resolution Matters: Real-Time Object Detection Latency
}

\author{

\IEEEauthorblockN{Qingyang Zhang}
\IEEEauthorblockA{\textit{Department of Computer Science} \\
\textit{University of Tsukuba}\\
Tsukuba, Japan \\
zhang.qingyang@sd.cs.tsukuba.ac.jp}
\and
\IEEEauthorblockN{Fumio Machida}
\IEEEauthorblockA{\textit{Department of Computer Science} \\
\textit{\textit{University of Tsukuba}}\\
Tsukuba, Japan \\
machida@cs.tsukuba.ac.jp}
\and
\IEEEauthorblockN{Laura Carnevali}
\IEEEauthorblockA{\textit{Department of Information Engineering} \\
\textit{University of Florence}\\
Florence, Italy \\
laura.carnevali@unifi.it}

}

\maketitle

\begin{abstract}
We present an input-resolution dependent distribution model for latency in real-time object detection. 
\textcolor{black}{We model total latency as the convolution of preprocessing, inference, and postprocessing distributions under a simplifying independence approximation, with selected stage parameters expressed as functions of source-image resolution.}
Under this assumption, the probability density of the total latency is the convolution of the stage-wise densities, and its cumulative distribution function (CDF) provides the distribution of end-to-end detection time.
Each stage is modeled by a parametric distribution (e.g., Exponential, Erlang, Normal, Gamma), with parameters expressed as functions of the source-image resolution.
Experiments with YOLOv11n on NVIDIA Jetson Orin NX using COCO2017 images across multiple resolutions assess the proposed models against fixed-parameter baselines using Kolmogorov Smirnov, Anderson Darling, and Cramér von Mises statistics. 
The results indicate that resolution-aware parameterization can improve distributional approximation in the measured setting, particularly for the more flexible Normal and Gamma models, while the quality of fit remains distribution dependent.
Our contribution is a theoretically grounded and lightweight formulation for studying resolution-dependent latency distributions in a measured object detection pipeline.
\end{abstract}

\section{Introduction}

Object detection systems have seen rapid development in recent years, driven by advancements in deep learning models such as YOLO series~\cite{redmon2018yolov3, yolov5,yolov8_ultralytics}, Faster R-CNN~\cite{ren2016fasterrcnnrealtimeobject}, and SSD~\cite{Liu_2016}. 
These systems are widely deployed in applications with strict real-time constraints, including autonomous vehicles, UAVs, and industrial inspection \cite{mao20233d, guleria2025systematic}. While much of the literature has focused on detection accuracy, latency performance, especially its variability and predictability, remains underexplored in both research and practice \cite{shin2025cf, chen2024latency}.

Object detection latency is often reported by scalar metrics such as the mean inference time or the average FPS, occasionally supplemented with quantiles or standard deviation \cite{electronics10030279,lazarevich2023yolobench}. 
However, these metrics are only representative statistics and \textcolor{black}{provide limited} insight into the performance dynamics of real-time object detection processes.

The inference latency of modern \textcolor{black}{single-stage} object detectors typically comprises three processing stages: preprocessing (e.g., image resizing and normalization), neural network inference, and postprocessing (e.g., non-maximum suppression). 
Although most end-to-end benchmarks treat the pipeline as a black box, our observation, which is based on the internal structure of YOLOv11n, suggests that the preprocessing latency varies significantly with the size of input image. 
The inference and postprocessing stages run on the fixed-size internal representations, and thus the latency is largely independent of the input image size. 
Therefore, the precise latency model needs to consider the performance characteristics of internal stages that may depend on the input image size.

In this study, we propose a distribution function 
for real-time object detection latency
\textcolor{black}{that combines three stage-wise distributions under a simplifying independence approximation.} 
The proposed distribution function is parameterized \textcolor{black}{by} the input resolution performance characteristics are captured in the model. 
In particular, we consider resolution-dependent distribution parameters that can be estimated from the mean and variance of empirical data.

To evaluate the proposed performance model, we conduct empirical experiments on the NVIDIA Jetson Orin NX using the YOLOv11n model~\cite{yolo11_ultralytics}, with input images from the COCO2017 dataset across multiple resolutions. 
\textcolor{black}{Our results provide an empirical assessment of the proposed formulation in this measured setting and show where resolution-dependent parameterization improves the approximation over fixed-parameter baselines.}

\vspace{1em}
    
    

\noindent The novelty and contributions of this work are threefold:
\begin{itemize}
    \item \textbf{Theoretical latency model:} We formalize the total latency of real-time object detection as a composition of stage-wise probability distributions (preprocessing, inference, and postprocessing), providing a principled foundation for distribution-level analysis. 
    
    \item \textbf{Resolution-aware parameterization:} \textcolor{black}{We introduce resolution-dependent distribution functions by explicitly linking statistical parameters to input image size, allowing intermediate resolutions to be approximated from anchor-resolution measurements in the studied setting.}
    
    \item \textbf{Empirical assessment and practical insights:} \textcolor{black}{Through experiments with multiple resolutions, we compare the proposed model with fixed-parameter baselines under CDF-based goodness-of-fit statistics and identify both improvements and remaining limitations in stage-wise latency approximation.}
\end{itemize}

\vspace{1em}

The remainder of this paper is organized as follows.
Section~\ref{sec: Background} motivates the problem, Section~\ref{sec:performance_model} presents the proposed latency model, Section~\ref{sec: Experiment design} describes the experimental setup, Section~\ref{sec: Evaluation} reports the evaluation, Section~\ref{sec: Discussion} discusses implications and limitations, Section~\ref{sec: Related Work} summarizes related work, and Section~\ref{sec: conclusion and future work} concludes the paper.

\section{\textcolor{black}{Background}}
\label{sec: Background}

In real-time object detection systems, the total \textcolor{black}{detection} latency can be decomposed into three sequential stages: preprocessing, inference, and postprocessing. 
From a modeling perspective, the overall latency and the stage-wise latencies are of interest. For instance, besides the total latency, the performance of a specific stage (i.e., preprocessing) is crucial for bottleneck analysis~\cite{oh2025optimizing}. 
The stage-wise performance models are also necessary for fine-grained performance analysis~\cite{zhang2024performability}.

A common way to report detection latency is through average runtime or frames per second. However, such aggregate metrics cannot capture the stochastic behavior of runtime, which is especially important for resource-constrained edge platforms. 
Since input image size affects operations such as resizing, normalization, and memory movement, we first examine whether the total latency distribution changes with input resolution.

\begin{figure}[htbp]
    \centering
    \includegraphics[width=\linewidth]{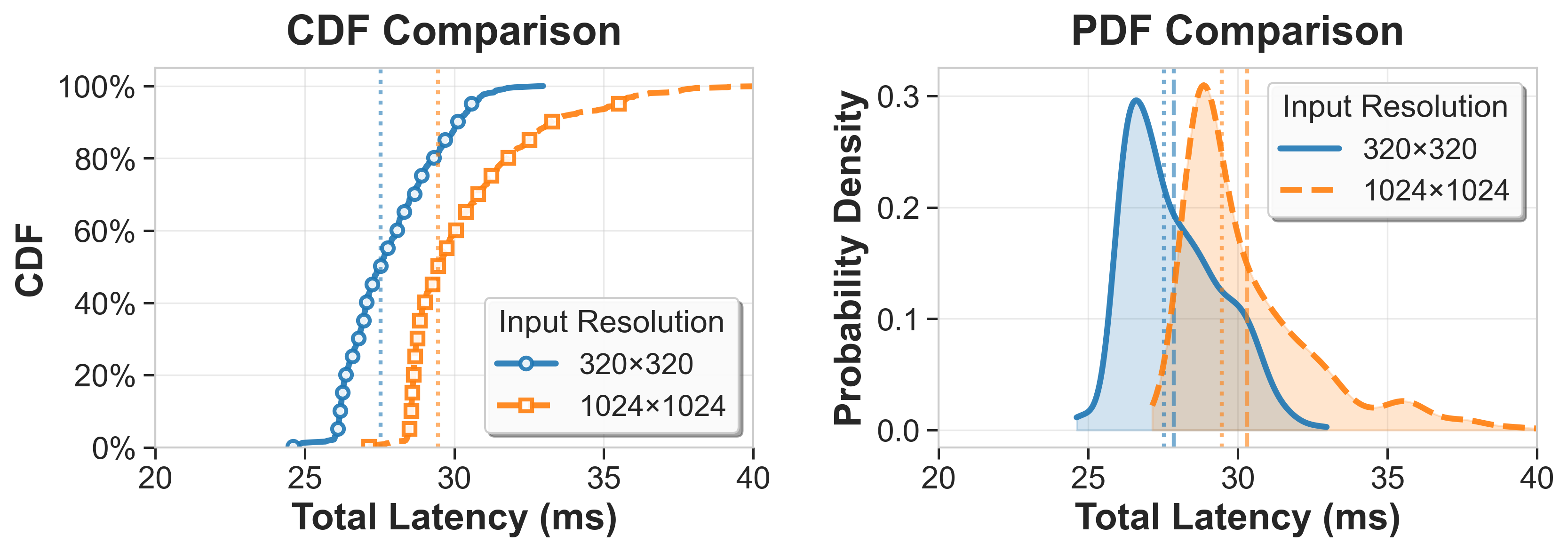}
    \caption{\textcolor{black}{Empirical CDF and PDF comparisons of total latency at the two anchor input resolutions on the Jetson platform.}}
    \label{fig:empirical_total_latency}
\end{figure}

As a motivating example, Figure~\ref{fig:empirical_total_latency} shows the empirical CDFs and PDFs of total detection latency at two anchor resolutions, $320 \times 320$ and $1024 \times 1024$. 
The curves indicate that input resolution affects both completion-time quantiles and the overall distribution shape. 
However, total-latency observations alone do not reveal which pipeline stage causes these differences.

\begin{figure}[htbp]
    \centering
    \includegraphics[width=\linewidth]{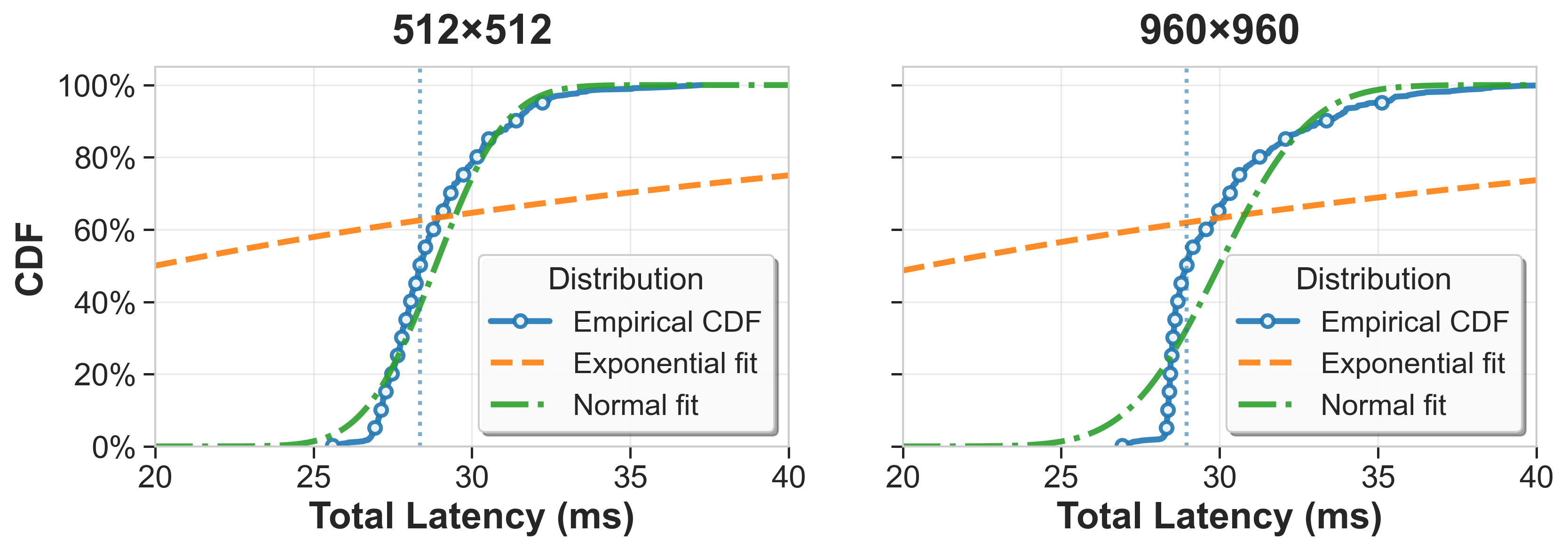}
    \caption{\textcolor{black}{Empirical total-latency CDFs and directly fitted Exponential and Normal CDFs at two intermediate input resolutions.}}
    \label{fig:fitted_total_latency_cdf}
\end{figure}

Figure~\ref{fig:fitted_total_latency_cdf} further compares empirical total-latency CDFs with directly fitted Exponential and Normal CDFs at $512 \times 512$ and $960 \times 960$. 
The comparison suggests that distribution choice matters: the Normal distribution follows the empirical CDF more closely, while the Exponential distribution provides a simpler but less accurate reference.

To interpret these observations, we consider the structure of YOLO-like detectors. 
After preprocessing, inference and postprocessing operate on fixed-size tensors due to model design and deployment optimizations, making their latency largely insensitive to the original image size~\cite{nvidia_cvcuda_2025,nvidia_orin_latency_2025}. 
\textcolor{black}{By contrast, framework preprocessing transforms the supplied source image into the fixed model tensor through spatial transformation, normalization, and channel permutation. The cost can therefore depend on the source-image dimensions~\cite{shahriar2020study,liu2024esod}.}
Therefore, preprocessing latency is expected to be both stochastic and resolution-dependent.

This motivates our modeling approach: instead of treating total latency as a black-box distribution, we decompose it into stage-wise components and introduce resolution-dependent parameters where appropriate. 
In the next section, we formally define the proposed performance model, including candidate distributions and parameterization strategies.

\section{Resolution-dependent latency distribution}
\label{sec:performance_model}

To accurately capture the variability of real-time object detection latency, we formulate the distribution function of the latency distribution with an input resolution parameter. 

\subsection{Object detection latency}
Let $T$ denote the total end-to-end latency of an object detection pipeline under an input image of resolution $r$. 
We are interested in its distribution function:
\begin{equation}
F(t, r) = \Pr(T \leq t \mid r),
\end{equation}
which represents the probability that the detection process finishes within $t$ milliseconds for resolution $r$.

We assume that the total latency $T$ can be decomposed into three stage-wise random variables:
\[
T = T_{\text{pre}} + T_{\text{infer}} + T_{\text{post}},
\]
corresponding to preprocessing, inference, and postprocessing stages, respectively. 
Let their probability density functions (pdfs) be denoted as $f_{\text{pre}}(x, r)$, $f_{\text{infer}}(y, r)$, and $f_{\text{post}}(z, r)$. 
\textcolor{black}{We adopt mutual independence among the three stage-wise latencies as a simplifying assumption for the convolution-based model.}
This assumption is evaluated through pairwise \textcolor{black}{correlations and two distribution-level impact metrics} in Section~\ref{sec:stage_dependence}.
Under this assumption, \textcolor{black}{for Exponential, Erlang, and Gamma, which have positive support, the total-latency pdf for $t\geq0$} is given by:
\begin{equation}
f_{T}(t, r) = \int_{0}^{t} \int_{0}^{t-x} 
f_{\text{pre}}(x, r) \, f_{\text{infer}}(y, r) \, f_{\text{post}}(t-x-y, r) 
\, dy \, dx.
\end{equation}

The corresponding CDF is obtained by integration:
\begin{equation}
F(t, r) = \int_{0}^{t} f_{T}(u, r)\, du.
\end{equation}


This formulation explicitly defines the total latency distribution as the three-fold convolution of stage-specific distributions. 
Conceptually, preprocessing handles raw input images, inference executes the neural network model, and postprocessing applies output transformations such as non-maximum suppression. 
The convolutional framework thus provides a \textcolor{black}{tractable way, under the stated independence approximation,} to capture their combined impact on end-to-end latency while retaining stage-wise interpretability.




\subsection{Resolution-Dependent Parameters}
We model each stage distribution $F_{s}(t, r)$ $(s\in {pre, inf, post})$ using one distribution.
Importantly, we generalize the parameters of these distributions to be functions of input resolution $r$. Formally,
\begin{equation}
F_{s}(t, r; \theta(r)),
\end{equation}
where $\theta(r)$ denotes the parameter set associated with the latency distribution of stage \textit{s}. 
If no correlation exists between resolution and a parameter, the function reduces to a constant, i.e., $\theta(r) = \theta$.

This formulation is based on three modeling assumptions:  
\begin{enumerate}
  \item Stage-wise latencies are independent random variables.
  \item Each stage latency follows a chosen parametric distribution family.
  \item Distribution parameters may depend on resolution $r$; if no dependency is observed, they are treated as constants.
\end{enumerate}

\textcolor{black}{These assumptions provide flexibility and interpretability. The convolutional framework gives a tractable way to combine stage-wise behaviors, while the parameterization $\theta(r)$ represents how input resolution may influence measured latency. The following sections empirically assess this approximation in the studied pipeline.}







\subsection{Candidate Distribution Families}
Four popular distribution families are considered for the stage-wise latency distributions. 
\textcolor{black}{The Exponential distribution is the Gamma special case with unit shape, whereas Erlang restricts the Gamma shape to a positive integer. We retain the separate names to emphasize these parameter constraints. Exponential, Erlang, and Gamma have positive support. The untruncated Normal candidate is defined on $\mathbb{R}$, although all measured latency samples are non-negative.}

\begin{itemize}
  \item \textbf{Exponential:}
  \[
  F_{s}(t, r) = 1 - e^{-\lambda(r) t}.
  \]

  \item \textbf{Erlang:}
  \[
  F_{s}(t, r) = 1 - \sum_{i=0}^{k-1} \frac{(\lambda(r) t)^i}{i!} e^{-\lambda(r) t}.
  \]

  \item \textbf{Normal:}
  \[
  F_{s}(t, r) = \Phi\!\left(\frac{t-\mu(r)}{\sigma(r)}\right),
  \]
  where $\Phi(\cdot)$ is the standard Gaussian CDF.

  \item \textbf{Gamma:}
  \[
  F_{s}(t, r) = \frac{1}{\Gamma(\alpha(r))} \, \gamma\!\left(\alpha(r), \, \beta(r) t\right),
  \]
  where $\alpha(r)$ is the shape parameter, $\beta(r)$ is the rate parameter, 
  $\Gamma(\cdot)$ is the Gamma function, and $\gamma(\cdot,\cdot)$ is the lower incomplete Gamma function.
\end{itemize}

Here, $\lambda(r)$, $\mu(r)$, $\sigma(r)$, $\alpha(r)$, and $\beta(r)$ denote resolution-dependent parameters. 
For example, if empirical evidence shows that $\lambda$ or $\beta$ scales approximately linearly with resolution $r$, they can be expressed as $\lambda(r) = a_{\lambda} r + b_{\lambda}$ or $\beta(r) = a_{\beta} r + b_{\beta}$.

The convolution of stage-wise latency distributions yields composite forms depending on the selected family. 
For the Normal distribution, the closure property of Gaussian variables ensures that the sum of three stage-wise Normal distributions remains Normal, with the mean and variance given by the sums of the respective stage parameters. 

\section{Experiment design}
\label{sec: Experiment design}

In this section, we will provide our experiment design and configuration, also evaluation metrics we plan to use,
since the validity of this formulation cannot be established purely at the theoretical level. 
In particular, it is necessary to evaluate (i) whether the preprocessing/inference/postprocessing stage indeed exhibits resolution-dependent behavior, (ii) whether the assumed distribution families adequately describe stage-wise latencies, and (iii) whether parameterizations such as linear resolution–parameter mappings generalize across unseen configurations. 

\subsection{Hardware and Software Configuration}

All experiments are conducted on an NVIDIA Jetson Orin NX 16GB embedded AI platform, a representative edge-computing device widely used in real-time vision applications. 
The software environment consists of Ubuntu 22.04.5 LTS, PyTorch 2.5.0, and torchvision 0.20.0. 
As the target detector, we employ the YOLOv11n model trained on the COCO2017 dataset~\cite{lin2015microsoft} for 100 epochs.

The test dataset consists of 560 images selected from the COCO2017 validation set. 
This subset is then uniformly resized \textcolor{black}{before the detector API call} to a range of square resolutions: $320 \times 320$, $416 \times 416$, $512 \times 512$, \textcolor{black}{$640 \times 640$,} $704 \times 704$, $960 \times 960$, and $1024 \times 1024$ for cross-resolution evaluation. 
For each resolution, the latency of the object detector is measured per image.


In this paper, we use the term \emph{resolution} to refer to the \textcolor{black}{externally prepared source-image size supplied to the detector}. 
Since all experiments are conducted on square images ($w = h$), we denote the resolution simply by the image width $r\textcolor{black}{\equiv r_s}$, which uniquely determines both dimensions and the total pixel count ($r^2$). 
This choice avoids ambiguity and simplifies parameterization in subsequent analysis.
\textcolor{black}{The detector retains its default internal model size $r_m=640$, so the experiment varies $r_s$ while holding the $640\times640$ neural-network tensor fixed.}
Among these resolutions, $r = 320$ and $r = 1024$ serve as anchor points for parameter estimation and baseline comparisons, 
while intermediate resolutions ($r = 416$, $512$, $704$, and $960$) are used to \textcolor{black}{assess how the proposed resolution-dependent model approximates latency distributions away from the anchor points}.
\textcolor{black}{The native $640\times640$ case is reported separately and excluded from the linear parameter estimation.}

\subsection{Latency Measurement Methodology}

The stage-wise latency of the YOLO detector was obtained directly from the 
official Ultralytics YOLO framework. For each input image, the framework 
internally records execution time for three distinct phases: 
\textit{pre-processing}, 
\textit{inference} (neural network forward pass), and 
\textit{post-processing} (e.g., non-maximum suppression). 
These values are automatically reported through the 
\texttt{results.speed} attribute of the YOLO API, which ensures a standardized and framework-consistent measurement of computational cost~\cite{ultralytics_speed_2025}.
\textcolor{black}{External image loading and resizing to $r_s\times r_s$ occur before the detector call and are excluded from the reported timings. Internal preprocessing timing begins after the prepared source image enters the framework.}
The latency values from YOLO’s internal profiler were then collected for each image and exported into CSV files. 
This allowed us to compute both per-image statistics and aggregated results across different resolutions, which were subsequently used for statistical fitting and distribution analysis.
We deliberately adopted YOLO’s internal profiler rather than relying on external timers (e.g., Python’s \texttt{time} or CUDA event wrappers)~\cite{ultralytics_issue_3700}. 

\subsection{Distribution Fitting Methodology}

To analyze the distributional characteristics of \textcolor{black}{stage-wise latency}, we adopt a parametric modeling approach using standard probability distributions: Exponential, Erlang, Normal and Gamma.

The core fitting process utilizes the latency data obtained at each resolution. For modeling and parameter estimation, we employ the \texttt{scipy.stats} package and apply the maximum likelihood estimation (MLE) method to obtain distribution parameters.

To construct an estimation model, we focus on two anchor resolutions and collect the fitted parameters at these two points. 
A linear regression is then performed over the observed values to approximate the trend across resolutions. 
\textcolor{black}{This regression-based model provides latency-distribution approximations for intermediate resolutions, which are then used for empirical assessment.}

\subsection{Baseline vs. Proposed model}
\textcolor{black}{To assess the proposed resolution-dependent distribution model, we compare it against three fixed-parameter baseline configurations:}

\begin{itemize}
  \item \textbf{Model 1:} A distribution fitted at $r=320$ and reused across all resolutions.
  \item \textbf{Model 2:} The distribution fitted at $r = 1024$ and reused across all resolutions. 
  \item \textbf{Model 3:} \textcolor{black}{A fixed, resolution-agnostic baseline estimated only from pooled stage-wise latency observations at the two anchor resolutions, $r=320$ and $r=1024$. For each stage, 1,500 samples are drawn with replacement from the 1,120 anchor observations. Maximum-likelihood parameters are computed for each draw and averaged over 30 bootstrap repetitions. The four intermediate resolutions are excluded from parameter estimation and used only for held-out evaluation.}
  \item \textbf{Proposed Model:} \textcolor{black}{A hybrid model that uses resolution-dependent parameter functions for resolution-sensitive stages and fixed two-anchor Model~3 parameters for the remaining stages.}
\end{itemize}

\textcolor{black}{None of the three baselines uses observations from the four held-out resolutions. Model~1 and Model~2 reuse parameters from one endpoint anchor. Model~3 pools observations from both anchors but discards their resolution labels, thereby providing a resolution-agnostic comparator with the same calibration resolutions as the proposed model.}

\textcolor{black}{When modeling total latency, all three stages use the same distribution family to retain a tractable convolution. The proposed hybrid mapping is distribution-specific. For the Exponential and Erlang families, preprocessing uses a resolution-dependent function, whereas inference and postprocessing use fixed Model~3 parameters. For the Normal and Gamma families, preprocessing and inference use resolution-dependent functions, whereas postprocessing uses fixed Model~3 parameters.}

In the following sections, we compare the baselines and proposed function against empirical latency data across multiple resolutions using goodness-of-fit metrics.

\subsection{Evaluation Metrics}

We assess the goodness-of-fit between the empirical latency distribution and the estimated (or directly fitted) distributions using standard statistical tests based on the CDF:

\begin{itemize}
  \item \textbf{Kolmogorov–Smirnov (KS) statistic:} 
  The KS test quantifies the maximum absolute distance between the empirical CDF and the fitted CDF. 
  It is highly interpretable: a smaller KS value means that the fitted distribution never deviates far from the empirical one at any point. 
  However, it only captures the single worst-case difference and may overlook more subtle discrepancies distributed across the domain.
  Its value lies in the interval $[0,1]$, where $0$ indicates a perfect fit.

  \item \textbf{Anderson–Darling (AD) statistic:} 
  The AD test is a refinement of CDF-based goodness-of-fit measures that assigns greater weight to the distribution tails. 
  This property makes it particularly relevant in latency modeling, where rare but large delays (tail events) can critically affect real-time system performance. 
  A lower AD value indicates that the fitted distribution better captures such extreme behaviors.
  AD takes values in $[0,\infty)$, with $0$ representing a perfect fit and larger values reflecting poorer agreement, especially in the tails.

  \item \textbf{Cramér–von Mises (CvM) statistic:} 
  The CvM test evaluates the integrated squared difference between the empirical CDF and the fitted CDF across the entire range. 
  Unlike KS (which focuses on the maximum deviation) or AD (which emphasizes the tails), CvM balances global accuracy by considering discrepancies throughout the distribution. 
  This makes it a useful complement for assessing overall fidelity of the fitted model.
  Its range is $[0,\infty)$, where smaller values indicate a better global fit.
\end{itemize}

Taken together, these three metrics provide a comprehensive evaluation: KS captures worst-case deviation, AD emphasizes tail accuracy, and CvM reflects overall fit quality. 
This multi-perspective assessment is essential for latency-sensitive applications, where both average-case accuracy and rare tail behaviors must be reliably modeled.

Notably, we do not employ probability density function (PDF)-based metrics (e.g., KL divergence) in the core evaluation, due to their sensitivity to histogram binning and instability across small sample sizes.

\section{Results}
\label{sec: Evaluation}
We first show the stage-wise latency trends and assess the stage-dependence approximation used in the total latency model. 
We then construct resolution-dependent parametric functions and evaluate the goodness-of-fit of the proposed model at both the stage level and the end-to-end total-latency level.

\subsection{Latency Trends Across Processing Stages}
\label{sec:stage_dependence}

We begin our analysis by measuring the average latency of YOLOv11n for each input image across the three stages. 
The results are summarized in Table~\ref{tab:stage_summary_latency}, with all values reported in milliseconds.
%
As the measurement results show, the inference stage dominates the absolute latency in all cases. 
The mean inference latency is nearly constant across resolutions (about 20.3--20.8 ms), which is caused by YOLOv11n processing fixed-size internal tensors. 
In contrast, preprocessing latency generally increases with input resolution, except for the native $640 \times 640$ case, \textcolor{black}{where the internal spatial transformation is reduced; the preprocessing stage itself is not skipped.}

\begin{table}[htbp]
\centering
\scriptsize
\caption{\textcolor{black}{Stage-wise latency summaries under different input resolutions. Each entry reports mean (STD) in ms.}}
\label{tab:stage_summary_latency}
\begin{tabular}{lrrr}
\hline
Resolution       & Preprocessing & Inference & Postprocessing \\ \hline
320x320   & 4.4341 (0.3734) & 20.2631 (0.9487) & 3.1556 (1.0049) \\
416x416   & 4.8196 (0.3853) & 20.6361 (1.1370) & 3.0920 (1.0164) \\
512x512   & 5.2795 (0.4742) & 20.5289 (1.2256) & 3.0445 (0.9226) \\
640x640   & 2.0464 (0.4623) & 20.7755 (1.4314) & 3.1548 (1.0670) \\
704x704   & 5.8610 (0.4335) & 20.2786 (1.3780) & 2.9083 (0.8159) \\
960x960   & 6.6876 (0.5383) & 20.3547 (1.6689) & 2.9399 (0.8296) \\
1024x1024 & 6.9465 (0.5728) & 20.3402 (1.5901) & 3.0282 (0.9639) \\ \hline
\end{tabular}
\end{table}

Table~\ref{tab:resolution_correlation_summary} further quantifies these trends using correlation analysis, excluding the native $640 \times 640$ case\textcolor{black}{.} The mean preprocessing latency has a strong positive correlation with \textcolor{black}{source-image} resolution ($r=0.9981$, $R^2=0.9963$), while the mean inference latency shows no statistically significant resolution dependence. 
When we look at the standard deviations (STDs), the preprocessing latency increases with resolution ($r=0.9275$), and the inference latency also grows despite its stable mean ($r=0.9783$). 
However, the postprocessing latency does not show \textcolor{black}{a} statistically significant trend. Therefore, we apply resolution-dependent fitting primarily to \textcolor{black}{the} preprocessing stage and additionally model inference variance for the Normal and Gamma distributions.

\begin{table}[h]
\scriptsize
\caption{\textcolor{black}{Resolution correlations of mean latency and latency STD by stage (excluding the $640$ width).}}
\label{tab:resolution_correlation_summary}
\begin{tabular}{llrrrl}
\hline
Statistic & Stage & Pearson $r$ & p-value & $R^2$ & Relationship \\ \hline
Mean & Preprocessing  & 0.9981  & 0.0000 & 0.9963 & Strong positive \\
Mean & Inference      & -0.3267 & 0.5274 & 0.1067 & Weak negative \\
Mean & Postprocessing & -0.6933 & 0.1266 & 0.4807 & Moderate negative \\
STD & Preprocessing  & 0.9275  & 0.0077 & 0.8602 & Strong positive \\
STD & Inference      & 0.9783  & 0.0007 & 0.9570 & Strong positive \\
STD & Postprocessing & -0.5455 & 0.2629 & 0.2976 & Moderate negative \\ \hline
\end{tabular}
\end{table}

\textcolor{black}{Table~\ref{tab:stage_correlation} assesses the independence approximation using 560 image-level observations per resolution. It reports pairwise Pearson correlations, the variance missed by ignoring cross-stage covariance, and the KS distance from totals obtained after independently shuffling the stage samples 500 times.}

\begin{table}[htbp]
\centering
\caption{\textcolor{black}{Pairwise stage correlations and the impact of the independence approximation. Variance inflation measures omitted covariance, and KS distance compares observed and independently recombined total-latency CDFs.}}
\label{tab:stage_correlation}
\scriptsize
\setlength{\tabcolsep}{3.2pt}
\resizebox{\columnwidth}{!}{%
\begin{tabular}{lrrrrr}
\hline
Resolution & Pre--Inf & Pre--Post & Inf--Post & \textcolor{black}{\shortstack{Variance\\inflation (\%)}} & \textcolor{black}{KS distance} \\ \hline
$320 \times 320$   & 0.088 & -0.074 & 0.144 & \textcolor{black}{13.7} & \textcolor{black}{0.074} \\
$416 \times 416$   & 0.151 & -0.040 & 0.077 & \textcolor{black}{11.3} & \textcolor{black}{0.039} \\
$512 \times 512$   & 0.197 & 0.069  & 0.095 & \textcolor{black}{19.5} & \textcolor{black}{0.061} \\
\textcolor{black}{$640 \times 640$} & \textcolor{black}{0.083} & \textcolor{black}{-0.037} & \textcolor{black}{0.090} & \textcolor{black}{10.2} & \textcolor{black}{0.046} \\
$704 \times 704$   & 0.205 & 0.120  & 0.251 & \textcolor{black}{32.5} & \textcolor{black}{0.137} \\
$960 \times 960$   & 0.241 & 0.178  & 0.264 & \textcolor{black}{35.2} & \textcolor{black}{0.137} \\
$1024 \times 1024$ & 0.368 & 0.030  & 0.142 & \textcolor{black}{30.0} & \textcolor{black}{0.121} \\ \hline
\end{tabular}
}
\end{table}

\textcolor{black}{Pairwise correlations are weak to moderate, but ignoring covariance underestimates total-latency variance by 10.2\%--35.2\%; the KS distance ranges from 0.039 to 0.137. Spearman and permutation checks reject strict independence, so we retain it only as a tractable approximation with quantified error. The $640\times640$ case is used only for this diagnostic and remains excluded from linear parameter fitting.}

\subsection{Fitted Parametric Functions}



Based on the latency data collected at the lowest ($r = 320$) and highest ($r = 1024$) input sizes, we derive analytical expressions that model the distribution parameters as functions of the image width $r$. 
Direct fits across the measured resolutions show clear resolution trends for several preprocessing parameters and for selected inference-variance parameters, so we use linear parameter functions as a lightweight approximation rather than as a universal assumption.
We construct three types of distributions to capture the latency characteristics of the preprocessing stage:

\begin{itemize}
  \item \textbf{Exponential Distribution}:
  \[
  \lambda(r) = -0.0001 \cdot r + 0.2625
  \]

  \item \textbf{Erlang Distribution}:
  \[
  \lambda(r) = -0.0012 \cdot r + 2.6258
  \]

  \item \textbf{Normal Distribution}:
  
  Preprocessing:
  \[
  \mu(r) = 0.0036 \cdot r + 3.2920, \quad
  \sigma(r) = 0.0003 \cdot r + 0.3016
  \]
  Inference:
  \[
  \mu(r) = 0.0001 \cdot r + 20.2280, \quad
  \sigma(r) = \textcolor{black}{0.0009} \cdot r + 0.6564
  \]

  \item \textbf{Gamma Distribution}:
  
  Preprocessing:
  \[
  \alpha(r) = -0.0016 \cdot r + 172.01, \quad
  \beta(r) = -0.0202  \cdot r + 45.165
  \]
  Inference:
  \[
  \alpha(r) = -0.4067 \cdot r + 599.44, \quad
  \beta(r) = -0.0201 \cdot r + 29.597
  \]
\end{itemize}

\textcolor{black}{The resolution-dependent functions are estimated from stage-level fits at the two anchor resolutions. Fixed Model~3 components are estimated separately by bootstrapping pooled observations from those same anchors, as defined in Section~\ref{sec: Experiment design}.}
\textcolor{black}{For the Erlang distribution, we fix the shape at $k=10$.}
Each function reflects the expected behavior of preprocessing latency as image size increases. 
For example, in both the Exponential and Erlang models, the rate parameter $\lambda$ decreases with increasing resolution, reflecting longer processing times due to greater image size. 
Similarly, in the Normal distribution, both the mean $\mu$ and standard deviation $\sigma$ exhibit a positive correlation with resolution, indicating not only a rise in average latency but also growing variance. \textcolor{black}{The Gamma preprocessing model analogously varies both its shape $\alpha$ and rate $\beta$.}

These parametric functions are then used to generate estimated latency distributions, which are evaluated in the following subsection through goodness-of-fit metrics against empirical distributions at intermediate resolutions.

\subsection{Goodness-of-Fit Evaluation}
\subsubsection{Stage-Level Goodness-of-Fit}
First, we evaluate the goodness-of-fit for individual stage latency functions. Table~\ref{tab:stage_level_ks_summary} summarizes representative KS statistics from the stage-level fitting results. We report KS because it directly measures the maximum CDF deviation. \textcolor{black}{Table~\ref{tab:total_latency_summary} additionally reports KS, CvM, and AD statistics for end-to-end latency.}
\textcolor{black}{The proposed Model~3 entries are recomputed against the same held-out stage traces. The proposed model uses resolution-dependent stage functions, whereas Model~3 uses fixed parameters estimated from the two anchors.}

\begin{table*}[htbp]
\centering
\small
\caption{Representative stage-level KS statistics for proposed and baseline models. Lower values indicate better CDF agreement. Ranges are over the four intermediate resolutions $416$, $512$, $704$, and $960$. \textcolor{black}{The lowest value in each scalar row is bold; for range rows, bold identifies the lowest mean KS over these resolutions.}}
\label{tab:stage_level_ks_summary}
\begin{tabular}{ll l rrrr}
\hline
Latency stage & Fitted distribution & Resolution(s) & Proposed & Model~1 & Model~2 & Model~3 \\ \hline
Preprocessing & Exponential & range & 0.632--0.654 & 0.639--0.759 & \textcolor{black}{\textbf{0.478--0.597}} & \textcolor{black}{0.548--0.670} \\
Preprocessing & Erlang & range & \textcolor{black}{\textbf{0.452--0.526}} & 0.568--0.901 & 0.424--0.694 & \textcolor{black}{0.356--0.668} \\
Preprocessing & Normal & $512\times512$ & \textcolor{black}{\textbf{0.298}} & 0.876 & 0.877 & \textcolor{black}{0.396} \\
Preprocessing & Normal & $960\times960$ & \textcolor{black}{\textbf{0.365}} & 1.000 & 0.513 & \textcolor{black}{0.678} \\
Preprocessing & Gamma & $512\times512$ & 0.491 & 0.905 & 0.890 & \textcolor{black}{\textbf{0.366}} \\
Preprocessing & Gamma & $960\times960$ & \textcolor{black}{\textbf{0.301}} & 1.000 & 0.521 & \textcolor{black}{0.700} \\
Inference & Normal & $512\times512$ & 0.158 & \textcolor{black}{\textbf{0.122}} & 0.228 & \textcolor{black}{0.193} \\
Inference & Normal & $960\times960$ & \textcolor{black}{\textbf{0.208}} & 0.257 & 0.215 & \textcolor{black}{0.216} \\
Inference & Gamma & $512\times512$ & 0.138 & \textcolor{black}{\textbf{0.117}} & 0.218 & \textcolor{black}{0.153} \\
Inference & Gamma & $960\times960$ & \textcolor{black}{\textbf{0.205}} & 0.256 & 0.207 & \textcolor{black}{0.258} \\ \hline
\end{tabular}
\end{table*}

\paragraph{Preprocessing Stage}
As shown in Table~\ref{tab:stage_level_ks_summary}, the preprocessing stage shows the clearest benefit from resolution-aware parameterization because its mean and variance both change with input size. The one-parameter Exponential distribution is useful mainly as a simple reference: the proposed function is comparable to fixed baselines but is not uniformly better, and Model~2 has lower KS values across the summarized range. The Erlang model improves over Model~1 in the summarized range, but Model~2 and Model~3 remain competitive under $704\times704$.

The strongest evidence appears for the more flexible Normal and Gamma distributions. Table~\ref{tab:stage_level_ks_summary} shows that preprocessing latency fitted by the Normal distribution has substantially lower KS values under the proposed model than under fixed endpoint baselines at representative intermediate and high resolutions. Preprocessing latency fitted by the Gamma distribution is also competitive, although \textcolor{black}{the two-anchor} Model~3 can be better at $512\times512$, indicating that pooled sampling can occasionally approximate intermediate settings well. Overall, the preprocessing results in Table~\ref{tab:stage_level_ks_summary} support the use of resolution-dependent parameters, especially for distribution families with enough flexibility to represent shifts in both scale and shape.

\paragraph{Inference Stage}
Inference-stage gains are smaller and less consistent because the inference mean latency is nearly resolution-invariant, as shown in Table~\ref{tab:stage_summary_latency} and Table~\ref{tab:resolution_correlation_summary}. The fitted parameter trends therefore mainly affect dispersion or distribution shape rather than the central location of the latency distribution. As a result, a fixed-parameter model fitted at one endpoint can already be close to the empirical inference distribution at some intermediate resolutions, and resolution-dependent parameterization does not always reduce the maximum CDF deviation. In Table~\ref{tab:stage_level_ks_summary}, inference latency fitted by the Normal distribution at $512\times512$ is better captured by Model~1 than by the proposed model, while inference latency fitted by the Gamma distribution at $960\times960$ shows a moderate improvement over Model~1 and remains close to Model~2. Thus, inference-stage parameterization should be interpreted as a secondary adjustment to variance and distribution shape rather than the main source of improvement.

\subsubsection{End-to-End Total Latency Goodness-of-Fit}

\begin{table*}[htbp]
\centering
\small
\caption{Compact total-latency goodness-of-fit summary. Values are means over the four intermediate resolutions; lower values are better. \textcolor{black}{The lowest value for each distribution and metric is bold.}}
\label{tab:total_latency_summary}
\setlength{\tabcolsep}{4.5pt} 
\begin{tabular}{l ccc ccc ccc ccc}
\hline 
Fitted distribution & \multicolumn{3}{c}{Proposed} & \multicolumn{3}{c}{Model 1} & \multicolumn{3}{c}{Model 2} & \multicolumn{3}{c}{Model 3} \\

 & KS & CvM & AD & KS & CvM & AD & KS & CvM & AD & KS & CvM & AD \\ \hline
Exponential & \textcolor{black}{0.573} & \textcolor{black}{46.82} & \textcolor{black}{215.47} & 0.585 & 49.48 & 226.26 & \textcolor{black}{\textbf{0.532}} & \textcolor{black}{\textbf{41.96}} & \textcolor{black}{\textbf{196.22}} & \textcolor{black}{0.560} & \textcolor{black}{45.00} & \textcolor{black}{208.17} \\
Erlang   & \textcolor{black}{0.415} & \textcolor{black}{24.29} & \textcolor{black}{122.20} & 0.478 & 30.59 & 145.22 & \textcolor{black}{\textbf{0.353}} & \textcolor{black}{\textbf{23.87}} & 121.27 & \textcolor{black}{0.404} & \textcolor{black}{24.03} & \textcolor{black}{\textbf{121.20}} \\
Normal      & \textcolor{black}{\textbf{0.166}} & \textcolor{black}{\textbf{5.57}}  & \textcolor{black}{\textbf{34.42}}  & 0.338 & 26.59 & 183.50 & 0.430 & 37.41 & 179.06 & \textcolor{black}{0.228} & \textcolor{black}{7.97}  & \textcolor{black}{44.19} \\
Gamma       & \textcolor{black}{\textbf{0.158}} & \textcolor{black}{\textbf{4.92}}  & \textcolor{black}{\textbf{35.28}}  & 0.348 & 30.13 & 195.63 & 0.385 & 37.69 & 209.99 & \textcolor{black}{0.225} & \textcolor{black}{8.63}  & \textcolor{black}{45.61} \\ \hline
\end{tabular}
\end{table*}

We obtain the total latency distribution of the object detection pipeline as the convolution of the stage-wise latency distributions.
Table~\ref{tab:total_latency_summary} summarizes the end-to-end total-latency comparison using the mean KS, CvM, and AD statistics over the four intermediate resolutions. 
\textcolor{black}{For the proposed model, Exponential and Erlang use a resolution-dependent preprocessing component with fixed two-anchor Model~3 inference and postprocessing components. Normal and Gamma use resolution-dependent preprocessing and inference components with a fixed two-anchor Model~3 postprocessing component. By contrast, each baseline uses fixed parameters for all three stages. Model~3 estimates those fixed parameters using only the two anchors.}


The proposed resolution-aware distribution function is most beneficial when the selected distributions can express the observed shift and spread of the total latency. 
%
\textcolor{black}{Under the Normal and Gamma families, the proposed model improves all three mean statistics relative to Model~1, Model~2, and the two-anchor Model~3.}

On the other hand, Exponential and Erlang distributions are less favorable because their limited parameterization cannot fully represent the measured total-latency shape. 
\textcolor{black}{Under Exponential and Erlang, Model~2 and Model~3 give slightly lower mean statistics than the proposed model.}
Nevertheless, the proposed model remains useful because it provides a single resolution-aware analytical construction that improves the end-to-end fit for the more expressive distributions and avoids choosing a fixed endpoint model whose accuracy depends strongly on the target resolution.

\newcommand{\appendixTotalBaseOneTable}{%
\begin{table}[htbp]
\centering
\caption{Goodness-of-fit comparison between proposed and Model~1 ($320\times320$) under different distributions}
\label{tab:fit_comparison_base1}
\begin{tabular}{l|ccc|ccc}
\hline
\multirow{2}{*}{Resolution} & \multicolumn{3}{c|}{\textbf{Proposed + Model 1}} & \multicolumn{3}{c}{\textbf{Model 1}} \\
 & KS & CvM & AD & KS & CvM & AD \\ \hline
\multicolumn{7}{c}{\textit{Exponential}} \\ \hline
416$\times$416 & 0.57 & 47.88 & 220.04 & 0.57 & 48.21 & 221.35 \\
512$\times$512 & 0.58 & 47.76 & 219.40 & 0.58 & 48.82 & 223.73 \\
704$\times$704 & 0.57 & 46.95 & 215.97 & 0.59 & 49.65 & 226.97 \\
960$\times$960 & 0.57 & 45.49 & 209.64 & 0.60 & 51.22 & 233.01 \\ \hline
\multicolumn{7}{c}{\textit{Erlang-10 stages}} \\ \hline
416$\times$416 & 0.42 & 25.65 & 128.24 & 0.43 & 27.07 & 133.42 \\
512$\times$512 & 0.42 & 25.21 & 126.00 & 0.45 & 28.39 & 137.57 \\
704$\times$704 & 0.42 & 24.71 & 124.01 & 0.49 & 30.89 & 146.38 \\
960$\times$960 & 0.39 & 21.44 & 109.87 & 0.54 & 36.00 & 163.53 \\ \hline
\multicolumn{7}{c}{\textit{Normal}} \\ \hline
416$\times$416 & 0.10 & 1.12 & 12.94 & 0.15 & 6.29 & 47.37 \\
512$\times$512 & 0.11 & 1.31 & 14.67 & 0.22 & 13.67 & 99.41 \\
704$\times$704 & 0.20 & 9.42 & 49.77 & 0.39 & 22.18 & 143.05 \\
960$\times$960 & 0.26 & 12.19 & 64.19 & 0.59 & 64.23 & 444.19 \\ \hline
\multicolumn{7}{c}{\textit{Gamma}} \\ \hline
416$\times$416 & 0.10 & 2.68 & 23.21 & 0.16 & 7.76 & 54.15 \\
512$\times$512 & 0.12 & 2.75 & 27.76 & 0.22 & 16.11 & 110.16 \\
704$\times$704 & 0.16 & 4.14 & 27.92 & 0.40 & 25.71 & 156.79 \\
960$\times$960 & 0.25 & 9.85 & 55.34 & 0.61 & 70.92 & 461.40 \\ \hline
\end{tabular}
\end{table}
}

\newcommand{\appendixTotalBaseTwoTable}{%
\begin{table}[htbp]
\centering
\caption{Goodness-of-fit comparison between proposed and Model~2 (parameters at $1024\times1024$) under different distributions}
\label{tab:fit_comparison_base2}
\begin{tabular}{l|ccc|ccc}
\hline
\multirow{2}{*}{Resolution} & \multicolumn{3}{c|}{\textbf{Proposed + Model 2}} & \multicolumn{3}{c}{\textbf{Model 2}} \\
 & KS & CvM & AD & KS & CvM & AD \\ \hline
\multicolumn{7}{c}{\textit{Exponential}} \\ \hline
416$\times$416 & 0.57 & 48.11 & 220.95 & 0.52 & 41.23 & 193.68 \\
512$\times$512 & 0.58 & 47.98 & 220.32 & 0.52 & 41.53 & 194.71 \\
704$\times$704 & 0.58 & 47.16 & 216.85 & 0.54 & 42.20 & 197.19 \\
960$\times$960 & 0.57 & 45.70 & 210.50 & 0.55 & 42.87 & 199.30 \\ \hline
\multicolumn{7}{c}{\textit{Erlang-10 stages}} \\ \hline
416$\times$416 & 0.42 & 25.94 & 129.35 & 0.35 & 25.64 & 128.90 \\
512$\times$512 & 0.42 & 25.51 & 127.12 & 0.32 & 23.86 & 121.91 \\
704$\times$704 & 0.43 & 24.91 & 124.75 & 0.35 & 24.64 & 124.62 \\
960$\times$960 & 0.40 & 21.58 & 110.40 & 0.39 & 21.34 & 109.66 \\ \hline
\multicolumn{7}{c}{\textit{Normal}} \\ \hline
416$\times$416 & 0.11 & 2.23 & 21.27 & 0.44 & 38.30 & 182.65 \\
512$\times$512 & 0.12 & 1.72 & 19.89 & 0.42 & 37.72 & 180.34 \\
704$\times$704 & 0.18 & 7.15 & 40.22 & 0.42 & 37.79 & 180.56 \\
960$\times$960 & 0.24 & 10.04 & 55.45 & 0.44 & 35.85 & 172.69 \\ \hline
\multicolumn{7}{c}{\textit{Gamma}} \\ \hline
416$\times$416 & 0.13 & 5.00 & 39.67 & 0.46 & 54.42 & 314.28 \\
512$\times$512 & 0.13 & 4.79 & 43.76 & 0.41 & 41.81 & 232.32 \\
704$\times$704 & 0.17 & 3.27 & 26.47 & 0.39 & 40.33 & 218.86 \\
960$\times$960 & 0.22 & 8.03 & 49.32 & 0.28 & 14.21 & 74.52 \\ \hline
\end{tabular}
\end{table}
}

\newcommand{\appendixTotalBaseThreeTable}{%
\begin{table}[htbp]
\centering
\caption{Goodness-of-fit comparison between proposed and Model~3 (random-sampling averaged parameters) under different distributions}
\label{tab:fit_comparison_base3}
\begin{tabular}{l|ccc|ccc}
\hline
\multirow{2}{*}{Resolution} & \multicolumn{3}{c|}{\textbf{Proposed + Model 3}} & \multicolumn{3}{c}{\textbf{Model 3}} \\
 & KS & CvM & AD & KS & CvM & AD \\ \hline
\multicolumn{7}{c}{\textit{Exponential}} \\ \hline
416$\times$416 & 0.57 & 47.95 & 220.30 & 0.69 &	78.23 &	359.90 \\
512$\times$512 & 0.58 & 47.82 & 219.66 & 0.67 &	76.70 &	352.17 \\
704$\times$704 & 0.57 & 47.01 & 216.24 & 0.66 &	76.02 &	349.21 \\
960$\times$960 & 0.57 & 45.56 & 209.92 & 0.63 &	71.12 &	325.08 \\ \hline
\multicolumn{7}{c}{\textit{Erlang-10 stages }} \\ \hline
416$\times$416 & 0.42 & 25.74 & 128.64 & 0.98 &	184.84 & 2815.85 \\
512$\times$512 & 0.42 & 25.30 & 126.40 & 0.97 &	184.51 & 2736.46 \\
704$\times$704 & 0.42 & 24.80 & 124.40 & 0.96 &	184.03 & 2700.36 \\
960$\times$960 & 0.40 & 21.53 & 110.27 & 0.95 &	182.45 & 2460.45 \\ \hline
\multicolumn{7}{c}{\textit{Normal}} \\ \hline
416$\times$416 & 0.11 & 2.27 & 22.27 & 0.22 & 10.42 & 48.44 \\
512$\times$512 & 0.12 & 1.71 & 20.59 & 0.17 & 5.74 & 28.00 \\
704$\times$704 & 0.18 & 7.10 & 40.17 & 0.18 & 6.58 & 37.29 \\
960$\times$960 & 0.24 & 10.04 & 55.85 & 0.30 & 8.39 & 63.11 \\ \hline
\multicolumn{7}{c}{\textit{Gamma}} \\ \hline
416$\times$416 & 0.13 & 4.95 & 40.00 & 0.25 & 13.18 & 61.51 \\
512$\times$512 & 0.13 & 4.73 & 44.13 & 0.20 & 7.38 & 34.35 \\
704$\times$704 & 0.17 & 3.28 & 26.70 & 0.18 & 7.66 & 40.74 \\
960$\times$960 & 0.23 & 8.11 & 49.90 & 0.28 & 6.77 & 53.22 \\ \hline
\end{tabular}
\end{table}
}

\section{Discussion}
\label{sec: Discussion}

The results indicate that the value of resolution-aware modeling depends on where resolution affects the pipeline and how much flexibility the selected distribution family provides. 
At the stage level, the clearest gains appear for preprocessing latency, whose mean and variance both change with input size.
By contrast, the inference latency has a nearly resolution-invariant mean. 
As a result, the resolution parameterization mainly adjusts the variance and shape of the fitted distribution. 
At the end-to-end level, the proposed model is most effective for the Normal and Gamma distributions, where resolution-dependent stage parameters can capture both shift and spread in the total-latency distribution. 
\textcolor{black}{The proposed model outperforms all three fixed baselines under Normal and Gamma, whereas Model~2 and Model~3 are slightly better under Exponential and Erlang.}
Although the Exponential and Erlang distributions are useful references, the resolution parameterization limits the achievable improvement and can make endpoint baselines competitive in some settings.

The main advantage of the proposed latency model is the analytically tractable \textcolor{black}{white-box} model that better fits the measured data with the resolution parameter.  
The total latency is represented as a convolution of stage-wise distributions, and each resolution effect is exposed through explicit parameter functions such as $\lambda(r)$, $\mu(r)$, $\sigma(r)$, $\alpha(r)$, and $\beta(r)$. 
This makes the model inspectable: users can check which stage contributes to the resolution sensitivity, which distributional parameter changes, and how the end-to-end distribution is assembled.
The proposed model can effectively \textcolor{black}{connect} empirical latency measurements with interpretable system structure.

\textcolor{black}{The current study is limited to one detector, one embedded platform, and square input resolutions. Extending the approach to non-square inputs may require parameterizing resolution by total pixels or by separate width and height terms, and different detector architectures may require different stage-wise dependency assumptions. These issues are left for future work.}

\section{Related Work}
\label{sec: Related Work}
\textcolor{black}{Object detection systems are commonly evaluated by accuracy and aggregate speed metrics such as mAP, FPS, or mean inference time. Prior benchmarking studies compare detector families, embedded platforms, and resolution settings, showing that deployment performance depends on both model architecture and hardware configuration~\cite{yolo11_ultralytics,lazarevich2023yolobench,cantero2022benchmarking}. Other work has examined resolution-dependent trade-offs and preprocessing overheads in computer vision pipelines~\cite{shahriar2020study,tse2023measuring,halpern2019one}. These studies motivate resolution-aware evaluation, but they usually report scalar latency summaries rather than modeling the full latency distribution.}

\textcolor{black}{Latency modeling for ML systems has also been studied through profiling, queueing, simulation, and system-level optimization~\cite{crankshaw2018inferline,hohman2024model}. Such approaches are useful for capacity planning and scheduling, but they typically do not express object-detection latency as a resolution-dependent, stage-wise probability distribution. Our work complements this literature by modeling preprocessing, inference, and postprocessing latencies as parametric distributions whose parameters may depend on input resolution.}

\section{Conclusion and Future Work}
\label{sec: conclusion and future work}

In this study, we proposed a \textcolor{black}{resolution dependent} latency model that links input resolution to the statistical properties of preprocessing latency. 
\textcolor{black}{The model leverages parametric distributions and provides a lightweight and interpretable way to approximate latency distributions at intermediate resolutions from anchor-resolution measurements.}
Our experiments based on real-world measurements by YOLOv11 show that resolution-aware parameterization can improve over fixed-parameter baselines in several settings. In particular, the proposed model gives clear improvements for preprocessing latency fitted by the Normal distribution and for end-to-end total latency fitted by the Normal and Gamma distributions, where all three CDF-based metrics are reduced against endpoint baselines. 
\textcolor{black}{The end-to-end improvement also holds against the two-anchor Model~3 under Normal and Gamma, but no consistent advantage is observed under Exponential or Erlang.}

For future work, we plan to extend the framework to other architectures and tasks, incorporate additional system-level variables (e.g., batch size, GPU utilization), and explore tail-sensitive modeling for better handling of latency spikes. 
Ultimately, integrating this model into runtime decision engines could enable adaptive resolution or processing strategies under dynamic constraints.

\section*{Acknowledgment}
This work was supported by JST BOOST Grant Number JPMJBS2414, and partly supported by a grant from the Telecommunications Advancement Foundation.

\bibliographystyle{IEEEtran}
\bibliography{IEEEabrv,cite}

%
%

\end{document}